# The Information Bottleneck EM Algorithm

**Gal Elidan** and **Nir Friedman**
School of Computer Science & Engineering, Hebrew University
{*galel,nir*}*@cs.huji.ac.il*

## Abstract

Learning with hidden variables is a central challenge in probabilistic graphical models that has important implications for many real-life problems. The classical approach is using the *Expectation Maximization* (EM) algorithm. This algorithm, however, can get trapped in local maxima. In this paper we explore a new approach that is based on the *Information Bottleneck* principle. In this approach, we view the learning problem as a tradeoff between two information theoretic objectives. The first is to make the hidden variables uninformative about the identity of specific instances. The second is to make the hidden variables informative about the observed attributes. By exploring different tradeoffs between these two objectives, we can gradually converge on a high-scoring solution. As we show, the resulting, *Information Bottleneck Expectation Maximization* (IB-EM) algorithm, manages to find solutions that are superior to standard EM methods.

## 1 Introduction

In recent years there has been a great deal of research on learning graphical models in general, and Bayesian networks in particular, from data [14, 9]. A central challenge in learning graphical models is learning with *hidden* (or *latent*) variables. Hidden variables typically serve as a summarizing mechanism that "captures" information from some of the observed variables and "passes" this information to some other part the network. As such, hidden variables can simplify the network structure and consequently lead to better generalization. The classical approach to learning with hidden variables is based on the *Expectation Maximization* (EM) [3, 12] algorithm. This algorithm performs a greedy search of the likelihood surface and is proven to converge to a local stationary point (usually a local maximum). Unfortunately, in hard real-life learning problems, there are many local maxima that can trap EM in a poor solution. Attempts to address this problem use a variety of strategies (e.g., [4, 8, 11, 15]).

In this paper, we introduce a new approach to learning Bayesian networks with hidden variables. In this approach, we view the learning problem as a tradeoff between two information theoretic objectives. The first objective is to compress information about the training data. The second objective is to make the hidden variables informative about the observed attributes to ensure they preserve the *relevant* information. By exploring different tradeoffs between these two objectives, we gradually converge on a high-scoring solution.

Our approach builds on the *Information Bottleneck* framework of Tishby *et al* [17] and its multivariate extension [6]. This framework provides methods for constructing new variables **T** that are stochastic functions of one set of variables **Y** and at the same time provide information on another set of variables **X**. The intuition is that the new variables **T** capture the relevant aspects of **Y** that are informative about **X**. We show how to pose the learning problem within the multivariate Information Bottleneck framework and derive a target *Lagrangian* for the hidden variables. We then show that this Lagrangian is an extension of the Lagrangian formulation of EM of Neal and Hinton [13], with an additional regularization term. By controlling the strength of this regularization term with a *scale parameter*, we can explore a range of target functions. On the one end of the spectrum there is a trivial target where compression of the data is total and all relevant information is lost. On the other end is the target function of EM.

The continuity of target functions allows us to learn using a procedure that is motivated by the *deterministic annealing* approach [15]. We start with the optimum of the trivial target function and slowly change the scale parameter while tracking the solution at each step on the way. We present an alternative view of the optimization problem in the joint space of the model parameters and the scaling factor. This provides an appealing method for scanning the range of solutions as in *homotopy continuation* [19]. Our procedure automatically increases the effective cardinality of the hidden variables as it progresses. Thus, by stopping the procedure at an earlier stage, we can reach a solution that has better generalization properties.

We extend our Information Bottleneck EM (IB-EM) framework for multiple hidden variables and any Bayesian network structure. We further show that the relation of the Information Bottleneck and EM holds for this case and for *variational EM* [10]. Finally, we show that the IB-EM algorithm is effective in learning models for *hard* real-life problems, and is often superior to EM based methods.

## 2 Bayesian Networks

Consider a finite set $\mathcal{X} = \{X_1, \ldots, X_n\}$ of random variables. A *Bayesian network* (BN) is an annotated directed acyclic graph that encodes a joint probability distribution over $\mathcal{X}$. The nodes of the graphs correspond to the random variables. Each node is annotated with a conditional probability distribution (CPD) of the random variable given its parents $\mathbf{Pa}_i$ in the graph $G$. The joint distribution can then be written as $P(X_1, \ldots, X_n) = \prod_{i=1}^{n} P(X_i | \mathbf{Pa}_i)$. The graph $G$ represents independence properties that are assumed to hold in the underlying distribution: Each $X_i$ is independent of its non-descendants given its parents $\mathbf{Pa}_i$.

Suppose we have a network structure $G$ over $\mathcal{X}$. Given a training set $\mathcal{D} = \{\mathbf{x}[1], \ldots, \mathbf{x}[M]\}$ that consists of instances of $\mathbf{X} \subset \mathcal{X}$, we want to learn parameters for the network. In the *Maximum Likelihood* setting we want to maximize the log-likelihood function $\log P(\mathcal{D} \mid G, \theta) = \sum_m \log P(\mathbf{x}[m] \mid G, \theta)$. This function can be equivalently written as $\mathbf{E}_{\hat{P}}[\log P(\mathbf{X} \mid g, \theta)]$ where $\hat{P}$ is the empirical distribution (frequencies) in $\mathcal{D}$. In practice, we also add a prior $P(\theta)$ on parameters, and then try to maximize $\mathbf{E}_{\hat{P}}[\log P(\mathbf{X} \mid \theta)] + \frac{1}{N} \log P(\theta)$ which results in MAP estimation (the $\frac{1}{N}$ term compensate for rescaling of the likelihood in the expectation). These priors can be thought of as adding imaginary instances that are distributed according to a certain distribution (e.g., uniform) to the training data [9]. Consequently, from this point on we view priors as modifying the empirical distribution with additional instances, and then apply the maximum likelihood principle.

## 3 Multivariate Information Bottleneck

The *Information Bottleneck* method [17] is a general information theoretic clustering framework. Given a joint distribution $Q(X, Y)$ of two variables, it attempts to finds a *bottleneck* variable $T$ defined by a stochastic function $Q(T \mid Y)$. This variable compresses the information in $Y$ while preserving the information that is relevant to $X$. For example, the variable $T$ might be used to define a soft clustering on words appearing in documents while preserving the information relevant to the topic of these documents.

The multivariate extension of this framework [6] allows the definition of several cluster variables as well as the desired interactions between them and the observed variables. The desirable interactions are represented via two Bayesian networks, one called $G_{in}$, representing the required compression, and the other called $G_{out}$ representing the independencies that we are striving for between the bottleneck variables and the target variables. In Figure 1, $G_{in}$ specifies that we want to minimize the information between $Y$ and $T$ (that compresses $Y$) and $G_{out}$ specifies that we want $T$ to make $Y$ and the variables $X_i$s independent of each other. These two objectives are conflicting, and the tradeoff between them "squeezes" the information $Y$ contains about the $X_i$'s.

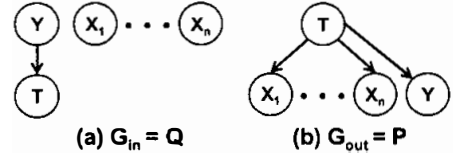

Figure 1: Definition of $G_{in}$ and $G_{out}$ for the Multivariate Information Bottleneck framework

Formally, the framework of Friedman *et al* [6], attempts to minimize the following objective function

$$\mathcal{L}[P, Q] = \mathbf{I}_Q(Y; T) + \gamma \mathbf{D}(Q(Y, T, \mathbf{X}) \| P(Y, T, \mathbf{X})) \quad (1)$$

where $\mathbf{D}$ is the Kulback Leibler divergence [2], $P$ and $Q$ are joint probabilities that can be represented by the networks of $G_{in}$ and $G_{out}$, respectively. Minimization is over possible parameterizations of $Q(T \mid Y)$ (the marginal $Q(Y, \mathbf{X})$ is given and fixed) and over possible parameterizations of $P(Y, T, \mathbf{X})$ that can be represented by $G_{out}$. In other words, we want to compress $Y$ in such a way that the distribution defined by $G_{in}$ is as close as possible to desired distribution of $G_{out}$. The scale parameter $\gamma$ balances these two factors. When $\gamma$ is zero we are only interested in compressing the variable $Y$ and we resort to the trivial solution of a single cluster (or an equivalent parameterization). When $\gamma$ is high we concentrate on choosing $Q(T \mid Y)$ that is close to a distribution satisfying the independencies encoded by $G_{out}$.

## 4 Learning a Hidden Variable

The main focus of the Information Bottleneck is on the learned distribution $Q(T \mid Y)$. This distribution can be thought of as a soft clustering of the original data. Our emphasis in this work is somewhat different. Given a dataset $\mathcal{D} = \{\mathbf{x}[1], \ldots, \mathbf{x}[M]\}$, we are interested in learning a better generative model describing the distribution of the observed attributes $\mathbf{X}$. That is, we want to give high probability to new data instances from the same source. In the learned network, the hidden variables will serve to "summarize" some part of the data while retaining the relevant information on some/all of the observed variables $\mathbf{X}$.

We start by extending the multivariate information bottleneck framework for the task of generalization where, in addition to the task clustering, we are also interested in learning the generative model of $P$. We first consider the case of a single hidden variable $T$, and extend the framework to several hidden variables in Section 5.

### 4.1 The Information Bottleneck EM Lagrangian

Intuitively, the task of generalization requires that we "forget" the specific training examples at hand but preserve the general form of the distribution over the observed variables. That is, we want to compress the identity of specific instances. On the other hand, since the observed variables are deterministically known when given the identity of a specific instance, we expect to loose information about the observed variables while performing this compression. This defines a tradeoff between the compression of the identity of specific instances and the preservation of the information relevant to the observed variables. We now formalize this idea.

We define a new variable $Y$ that denotes the instance identity. That is, $Y$ takes values in $\{1, \ldots, M\}$ and $Y[m] = m$. We define $Q(Y, \mathbf{X})$ to be the empirical distribution of the attributes $\mathbf{X}$ in the data, augmented with the distribution of the new variable $Y$. For each instance $y$, $\mathbf{x}[y]$ is the value $\mathbf{X}$ take in the specific instance. We now apply the Information Bottleneck with the graph $G_{in}$ of Figure 1. The choice of the graph $G_{out}$ depends on the network we want to learn. We take it to be the target Bayesian network with the additional variable $Y$ and $T$ as its parent. For simplicity, we consider the simple clustering model of $G_{out}$ where $T$ is the parent of $X_1, \ldots, X_n$. In practice, and as will be seen in section 6, any choice of $G_{out}$ can be used.

We can now apply the Information Bottleneck framework to this pair of graphs. This will attempt to define conditional probability $Q(T \mid Y)$ so that $Q(T, Y, \mathbf{X}) = Q(T \mid Y) Q(Y, \mathbf{X})$ can be approximated by a distribution that factorizes according to $G_{out}$. This construction will aim to find $T$ that captures the relevant information the instance identity has about the observed attributes.

We start by determining the objective function for the particular choice of $G_{in}$ and $G_{out}$ we are dealing with.

**Proposition 4.1:** *Minimizing the information bottleneck objective function Eq. (1) is equivalent to minimizing the Lagrangian*

$$\mathcal{L}_{EM} = \boldsymbol{I}_Q(T; Y) - \gamma \left( \boldsymbol{E}_Q[\log P(\mathbf{X}, T)] - \boldsymbol{E}_Q[\log Q(T)] \right)$$

*as a function of $Q(T \mid Y)$ and $P(\mathbf{X}, T)$.*

**Proof:** Using the chain rule and the structure of $P$, we can write $P(Y, \mathbf{X}, T) = P(Y \mid T) P(\mathbf{X}, T)$. Similarly, since $\mathbf{X}$ is independent of $T$ given $Y$, we can write $Q(Y, \mathbf{X}, T) = Q(Y \mid T) Q(T) Q(\mathbf{X} \mid Y)$. Thus,

$$\boldsymbol{D}(Q(Y, \mathbf{X}, T) \| P(Y, \mathbf{X}, T))$$
$$= \boldsymbol{E}_Q \left[ \log \frac{Q(Y \mid T) Q(T) Q(\mathbf{X} \mid Y)}{P(Y \mid T) P(\mathbf{X}, T)} \right]$$
$$= \boldsymbol{D}(Q(Y \mid T) \| P(Y \mid T)) + \boldsymbol{E}_Q[\log Q(\mathbf{X} \mid Y)] + \boldsymbol{E}_Q[\log Q(T)] - \boldsymbol{E}_Q[\log P(\mathbf{X}, T)]$$

By setting $P(Y \mid T) = Q(Y \mid T)$, the first term reaches zero, its minimal value. The second term is a constant (since we cannot change $Q(\mathbf{X} \mid Y)$). And thus, we need to minimize the last two terms and the result follows immediately. ∎

An immediate question is how this target function relates to standard maximum likelihood learning. To explore the connection, we use a formulation of EM introduced by Neal and Hinton [13]. Although EM is usually thought of in terms of changing the parameters of the target function $P$, Neal and Hinton show how to view it as a dual optimization of $P$ and an auxiliary distribution $Q$. Using the same notation, we can write the functional defined by Neal and Hinton as

$$\mathcal{F}[Q, P] = \boldsymbol{E}_Q[\log P(\mathbf{X}, T)] + \boldsymbol{H}_Q(T \mid Y)$$

where $\boldsymbol{H}_Q(T \mid Y) = \boldsymbol{E}_Q[-\log Q(T \mid Y)]$, and $Q(\mathbf{X})$ is fixed to be the observed empirical distribution.

**Theorem 4.2:** [13] *If $(Q*, P*)$ is a stationary point of $\mathcal{F}$, then $P^*$ is a stationary point of the log-likelihood function $\boldsymbol{E}_Q[\log P(\mathbf{X})]$.*

Moreover, Neal and Hinton show that an EM iteration corresponds to maximizing the choice $Q(T \mid Y)$ while holding $P$ fixed, and then maximizing $P$ while holding $Q(T \mid Y)$ fixed.

The form of $\mathcal{F}[Q, P]$ is quite similar to the IB-EM Lagrangian, and indeed we can relate the two.

**Proposition 4.3:** $\mathcal{L}_{EM} = (1 - \gamma) \boldsymbol{I}_Q(T; Y) - \gamma \mathcal{F}[Q, P]$

**Proof:** Using the identity $\boldsymbol{H}_Q(T \mid Y) = -\boldsymbol{E}_Q[\log Q(T)] - \boldsymbol{I}_Q(T; Y)$, we can write

$$\mathcal{F}[Q, P] = \boldsymbol{E}_Q[\log P(\mathbf{X}, T)] - \boldsymbol{E}_Q[\log Q(T)] - \boldsymbol{I}_Q(T; Y)$$

and the result follows immediately. ∎

As a consequence, minimizing the IB-EM Lagrangian is equivalent to maximizing the EM functional combined with an information theoretic regularization term. When $\gamma = 1$ the Lagrangian and the EM functional are equivalent and finding a local minima of $\mathcal{L}_{EM}$ is equivalent to finding a local maxima of the likelihood function.

### 4.2 The IB-EM Algorithm

For a specific value of $\gamma$, the Information Bottleneck EM (IB-EM) algorithm can then be described as iterations similar to the EM iterations of Neal and Hinton [13].

- E-step: Minimize $\mathcal{L}_{EM}$ by optimizing $Q(T \mid Y)$ while holding $P$ fixed.

- M-step: Minimize $\mathcal{L}_{EM}$ by optimizing $P$ while holding $Q$ fixed.

The M-Step is essentially the standard maximum likelihood optimization of Bayesian networks. To see that, note that the only term that involves $P$ is $\boldsymbol{E}_Q[\log P(\mathbf{X}, T)]$. This term has the form of a log-likelihood function with the "empirical" distribution $Q$. Since the distribution is over all the variables, we can use sufficient statistics of $P$ for efficient estimates. Thus, the $M$ step consists of computing expected sufficient statistics given $Q$, and then using a "plug-in" formula for choosing the parameters of $P$.

The E-step is a bit more involved. We need to optimize $Q(T \mid Y)$. To do this we use the following two results that follow from results of Friedman *et al* [6].

**Proposition 4.4:** $Q(T|Y)$ *is a stationary point of* $\mathcal{L}_{EM}$ *with respect to a fixed choice of* $P$ *if and only if for all values* $t$ *and* $y$ *of* $T$ *and* $Y$, *respectively,*

$$Q(t|y) = \frac{1}{Z(y)} Q(t)^{1-\gamma} \exp^{\gamma \mathbf{EP}(t,y)} \qquad (2)$$

*where* $\mathbf{EP}\,(t,y) \equiv \log P(\mathbf{x}[y],t)$ *and* $Z(y)$ *is a normalizing constant.*

**Proposition 4.5:** *A stationary point of* $\mathcal{L}_{EM}$ *is achieved by iteratively applying the self-consistent equations of Proposition 4.4.*

In most cases the stationary convergence point reached by applying these self-consistent equations will be a local maxima.

Combining this result, with the result of Neal and Hinton that show that optimization of $P$ increases $F(P,Q)$, we conclude that both the E-step and the M-step decrease $\mathcal{L}_{EM}$ until we reach stationary point.

### 4.3 Bypassing Local Maxima using Continuation

As discussed above, the parameter $\gamma$ balances the desire to compress the data and the desire to fit parameters to $G_{out}$. When $\gamma$ is close to 0, our only objective is compressing the data. and the effective dimensionality of $T$ will be 1 leading to a trivial solution. At larger values of $\gamma$ we pay more and more attention to the distribution of $G_{out}$, and we can expect additional states of $T$ to be utilized. Ultimately, we can expect each sample to be assigned to a different cluster (if the dimensionality of $T$ allows it). Proposition 4.3 tells us that at the limit of $\gamma = 1$ our solution will actually converge to one of the standard EM possible solutions.

Naively, we could allow a high cardinality for the hidden variable, set $\gamma$ to a "high" value and find the bottleneck solution at that point. There are several drawbacks to this approach. First, we will typically converge to a sub-optimal solution for a given cardinality and $\gamma$, all the more so for $\gamma = 1$ where there are many such maxima. Second, we often do not know the correct cardinality that should be assigned to the hidden variable. If we use a cardinality for $T$ that is too large, learning will be less robust and might become intractable. If $T$ has too low a dimensionality, we will not fully utilize the potential of the hidden variable. We would like to somehow identify the beneficial number of clusters without having to simply try many options.

To cope with this task, we adopt the *deterministic annealing* strategy [15]. In this strategy, we start with $\gamma = 0$ where a single cluster solution (high entropy) is optimal and compression is total. We then progress toward higher values of $\gamma$. This gradually introduces additional structure into the learned model.

There are several ways of executing this general strategy. The common approach is simply to increase $\gamma$ in fixed steps, and after each increment apply the iterative algorithm to re-attain a (local) minima with the new value of $\gamma$. On the problems we examine in Section 6, this naive approach did not prove successful. Instead, we use a more refined approach that utilizes *continuation methods* for executing this strategy. This approach provides means for automatically tuning the appropriate change in the size of $\gamma$, and also ensures we keep track of the solution from one iteration to the next.

To perform continuation, we view the optimization problem in the joint space of the parameters and $\gamma$. In this space we want to follow a smooth path from the trivial solution at $\gamma = 0$ to a solution at $\gamma = 1$. Furthermore we would like to require that the fix-point equations hold at all points along the path. Continuation theory [19] guarantees that, excluding degenerate cases, such a path, free of discontinuities, indeed exists.

We start by characterizing such paths. Note that once we fix the parameter $Q(T \mid Y)$, the parameters of the optimal choice of $P$ are determined as a function of $Q$. Thus, we take $Q(T \mid Y)$ and $\gamma$ as the only free parameters in our problem. As we have shown in Proposition 4.4, when the gradient of the Lagrangian is zero, Eq. (2) holds for each value of $t$ and $y$. Thus, we want to consider paths where all of these equations hold. We define

$$G_{t,y}(Q,\gamma) = -\log Q(t|y) + (1-\gamma)Q(t) + \gamma \mathbf{EP}\,(t,y) - \log Z(y) \qquad (3)$$

Clearly, $G_{t,y}(Q,\gamma) = 0$ exactly when Eq. (2) holds for $t$ and $y$. Our goal is then to follow an equi-potential path where all $G_{t,y}(Q,\gamma)$ functions are zero starting from some small value of $\gamma$ upto the desired EM solution at $\gamma = 1$.

Suppose we are at a point $(Q_0,\gamma_0)$ where all the $G_{t,y}$ functions are equal to 0. We want to move in a direction $\Delta = (dQ,d\gamma)$ so that $(Q_0 + dQ,\gamma_0 + d\gamma)$ also satisfies the fix-point equations. To do so, we want to find a direction $\Delta$, so that

$$\forall t,y, \;\; \nabla_{Q,\gamma} G_{t,y}(Q_0,\gamma_0) \cdot \Delta = 0 \qquad (4)$$

Computing the derivatives of $G_{t,y}(Q,\gamma)$ with respect to each of the parameters results in a derivative matrix $H(Q,\gamma)$. Rows of the matrix correspond to each of the $L = |T| \times |Y|$ functions of Eq. (3), and columns correspond to the $L$ parameters of $Q$ as well $\gamma$. The entries correspond to the partial derivative of the function associated with the row with respect to the parameter associated with the column.

To find a direction $\Delta$ that satisfies Eq. (4) we need satisfy the matrix equation

$$H(Q,\gamma)\Delta = 0 \qquad (5)$$

In other words, we are trying to find a vector in the nullspace of $H(Q,\gamma)$. Note that $H$ is an $L \times (L+1)$ ma-trix and the direction vector is of length $L + 1$. Thus, the null-space is of dimension $L + 1 - \text{Rank}(H(Q, \gamma))$. Numerically, excluding measure zero cases [19], we expect $\text{Rank}(H(Q, \gamma))$ to be full, i.e., $L$. Thus there is a unique (upto to scaling) solution to Eq. (5).

Finding this direction, however, can be costly. Notice that $H(Q, \gamma)$ is of size $L(L+1)$. This number is quadratic in the training set size, and so just computing the matrix is prohibitively expensive, even for small datasets. Instead, we resort to approximating $H(Q, \gamma)$ by a matrix that contains only the diagonal entries $\frac{\partial G_{t,y}(Q, \gamma)}{\partial Q(t|y)}$ and the last column $\frac{\partial G_{t,y}(Q, \gamma)}{\partial \gamma}$. In our case these are also the most significant values of the matrix since in the off-diagonal derivatives many terms are set to zero. Once we make the approximation, we can solve Eq. (5) in time linear in $L$.

Note that once we find a vector $\Delta$ that satisfies Eq. (5), we still need to decide on the size of the step we want to take in that direction. There are various standard approaches, such as normalizing the direction vector to a predetermined size. However, in our problem, we have a natural measure of progress. Recall that $I(T; Y)$ captures information about the samples. In all runs $I(T; Y)$ starts at 0, and is upper-bounded by the log of the cardinality of $T$. Moreover, the "interesting" steps in the learning process occur when $I(T; Y)$ grows. These are exactly the points where the balance between the two terms in the Lagrangian changes and the second term grows sufficiently to allow the first term to increase $I(T; Y)$.

With this intuition at hand, we want to normalize the step size by the expected change in $I(T; Y)$. If we are at a region where changes in the parameters are not influencing $I(T; Y)$, then we can make a big step. On the other hand, if the change has a strong influence on $I(T; Y)$, then we want to carefully track the solution. Formally, we compute $\nabla_{Q, \gamma} I(T; Y)$ and rescale the direction vector so that

$$(\nabla_{Q, \gamma} I_Q(T; Y))' \cdot \Delta = \epsilon$$

where $\epsilon$ is a predetermined step size. We also bound the minimal and maximal change in $\gamma$ so that we do not get trapped in too many steps or alternatively overlook the regions of change.

Finally, although the continuation method takes us in the correct direction, the approximation as well as inherent numerical instability can lead us to a suboptimal path. To cope with this situation, we adopt a commonly used heuristic used in *deterministic annealing*. At each value of $\gamma$, we slightly perturb the current solution and re-solve the self-consistent equations. If this perturbation leads to a higher Lagrangian, we take it as our current solution.

To summarize, our procedure works as follows: we start with $\gamma = 0$ for which only the trivial solution exists. At each stage we compute the direction of $\gamma$ and $Q(T|Y)$ that will leave the fix-point equations intact. We then take a small step in this direction and apply IB-EM iterations to attain the fix-point equilibrium at the new value of $\gamma$. We repeat these iterations until we reach $\gamma = 1$.

### 4.4 Regularization and Generalization

In the discussion so far we were concerned with reaching the value of $\gamma = 1$ with a good solution. As the experimental results below show, by using continuation methods our algorithm manages to reach solutions that are superior to running standard EM.

However, in many domains, maximizing the likelihood can lead to overfitting and poor generalization. Thus it is common in machine learning to use regularization to counterweight the tendency to overfit the training examples. We can use priors to reduce this effect. However, they do not guarantee the best generalization performance. As discussed above, we want to learn hidden variables with a large cardinality and use regularization to determine their effective cardinality

The information bottleneck formalization provides a form of regularization that arise naturally from the definition of the learning problem. When we learn with $\gamma < 1$, the compression term counteracts the tendency to overfit the data. Thus, we might get better generalization with parameters we estimate for intermediate values of $\gamma$. Indeed, when we examine models learned in different continuation iterations, we see that later iterations, when $\gamma$ is closer to 1, can degrade the generalization performance. This observation suggests that, for learning purposes, we would prefer to learn a model at some $\gamma^*$ that is usually smaller than 1. The technical challenge is how to select $\gamma^*$.

A relatively straightforward but somewhat costly approach is to use a cross-validation (CV) test. In this approach we perform $k$ runs of our algorithm, each one learning from $(k-1)/k$'th of the training data. In each run we evaluate intermediate models on the remaining $1/k$'th data. We can then estimate the generalization at different values of $\gamma$ by averaging the log-likelihood of the held-out data at this value of $\gamma$ in each of the $k$ folds. We use this evaluation to estimate $\gamma^*$, and then perform the continuation process on the whole training data up to the critical $\gamma^*$. It is important to stress that this approach utilizes *only* the training data in order to predict the value of $\gamma^*$ at which generalization will be best.

## 5 Multiple Hidden Variables

The framework we described in the previous section can easily accommodate learning networks with multiple hidden variables simply by treating $T$ as a vector of hidden variables. In this case, the distribution $Q(\mathbf{T} \mid Y)$ describes the *joint* distribution of the hidden variables for each value of $Y$, and $P(\mathbf{T}, \mathbf{X})$ describes their joint distribution with the attributes $\mathbf{X}$ in the desired network. Unfortunately, if the number of variables $\mathbf{T}$ is large, the representation of $Q(\mathbf{T} \mid Y)$ grows exponentially, and this approach becomes

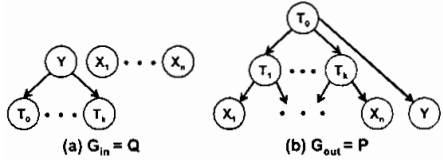

Figure 2: Definition of networks for the Multivariate Information Bottleneck framework with multiple hidden variables. (a) shows $G_{in}$ with the *mean field* assumption. (b) shows a possible hierarchy for $G_{out}$.

infeasible.

One strategy to alleviate this problem is to force $Q(\mathbf{T} \mid Y)$ to have a factorized form. This reduces the cost of representing $Q$ and also the cost of performing inference . As an example, we can require that $Q(\mathbf{T} \mid Y)$ is factored as a product $\prod_i Q(T_i \mid Y)$. This assumption is similar to the *mean field variational approximation* (e.g., [10]).

In the Multivariate Information Bottleneck framework, different factorizations of $Q(\mathbf{T} \mid Y)$ correspond to different choices of networks $G_{in}$. For example, the mean field factorization is achieved when $G_{in}$ is such that the only parent of each $T_i$ is $Y$, as in Figure 2. In general, we can consider other choices where we introduce edges between the different $T_i$'s. For any such choice of $G_{in}$, we get exactly the same Lagrangian as in the case of a single hidden variable. The main difference is that since $Q$ has a factorized form, we can decompose $\mathbf{I}_Q(\mathbf{T}; Y)$. For example, if we use the mean field factorization, we get $\mathbf{I}_Q(\mathbf{T}; Y) = \sum_i \mathbf{I}_Q(T_i; Y)$. Similarly, we can decompose $E_Q[\log P(\mathbf{X}, \mathbf{T})]$ into a sum of terms, one for each family in $P$. These two factorization can lead to tractable computation of the first two terms of the Lagrangian as written in Proposition 4.1. Unfortunately, the last term $E_Q[\log Q(\mathbf{T})]$ cannot be evaluated efficiently. Thus, we approximate this term as $\sum_i E_Q[\log Q(T_i)]$. For the mean field factorization, the resulting Lagrangian (with this approximation) has the form

$$\mathcal{L}_{EM}^{+} = \sum_i \mathbf{I}_Q(T_i; Y) - \gamma \left( E_Q[\log P(\mathbf{X}, T)] - \sum_i E_Q[\log Q(T_i)] \right)$$

As in the case of a single hidden variable, we can characterize fix-point equations that hold in stationary points of the Lagrangian.

**Proposition 5.1:** *Assuming a mean field approximation for $Q(\mathbf{T} \mid Y)$, a (local) maximum of $\mathcal{L}_{EM}^{+}$ is achieved by iteratively solving the following self-consistent equations for every hidden variable $i$ independently.*

$$Q(t_i|y) = \frac{1}{Z(i,y)} Q(t_i)^{1-\gamma} \exp^{\gamma \mathbf{EP}(t_i, y)}$$

*where* $\mathbf{EP}(t_i, y) \equiv E_{Q(\mathbf{T}|t_i, y)}[\log P(\mathbf{x}[y], \mathbf{T})]$ *and* $Z(i, y)$ *is a normalizing constant.*

The proof follows the lines of Theorem 7.1 of [6]. The only difference in the computation of the fix-point equations is in the derivative of $E_Q[\log P(\mathbf{x}[y], \mathbf{T})]$ with respect to $Q(t_i \mid y)$ resulting in the different form for $\mathbf{EP}(t_i, y)$, that can still be computed efficiently. It is easy to see that when a single hidden variable is considered, the two forms coincide.

A more interesting consequence of this discussion is that when $\gamma = 1$, maximizing $\mathcal{L}_{EM}^{+}$ is equivalent to performing *mean field EM* [10]. Thus, by using the modified Lagrangian we generalize this variational learning principle, and as we show below manage to reach better solutions. We can easily extend the same idea to describe a correspondence between different choices of $G_{in}$ and the "matching" structural approximation when applied to standard EM. For lack of space, we do not go into details here, although they are fairly straightforward.

To summarize, the IB-EM algorithm of section 4.2 can be easily generalized to handle multiple hidden variables by simply altering the form of $\mathbf{EP}(t_i, y)$ in the fix-point equations. All other details, such as the continuation method, remain unchanged.

# 6    Experimental Validation

To evaluate the IB-EM method, we examine its generalization performance on several types of models on three real-life datasets. In each architecture, we consider networks with hidden variables of different cardinality, where for simplicity we use the same cardinality for all hidden variables in the same network. We now briefly describe the datasets and the model architectures we use.

The Stock dataset records up/same/down daily changes of 20 major US technology stocks over a period of several years [1]. The training set includes 1213 samples and the test set includes 303 instances. We trained a Naive Bayes hidden variable model where the hidden variable is a parent of all the observations.

The Digits dataset contains 400 instances sampled from the USPS (US Postal Service) dataset of handwritten digits (see http://www.kernel-machines.org/data.html). This data includes 40 images for each digit. An image is represented by 256 variables, each denotes the gray level of one pixel in a $16 \times 16$ matrix. We discretized pixel values into 10 equal bins. We used 320 images as a training set and 80 as a test set.

On this data we tried several network architectures. The first is a Naive Bayes model with a single hidden variable. In addition, we examined more complex hierarchical models. In these models we introduce a hidden parents to each quadrant of the image recursively. The 3-level hierarchy has a hidden parent to each 8x8 quadrant, and then another hidden variable that is the parent of these four hidden variables. The 4-level hierarchy starts with 4x4 blocks, and has an additional intermediate level of hidden variables (total of 21 hidden variables).

The Yeast dataset contains expression measurement the baker's yeast genes in 173 experiments [7]. These experi-ments measure the yeast response to changes in its environmental conditions. For each experiment the expression of 6152 genes were measured. We discretized the expression of genes into ranges down/same/up by using a threshold of +/- standard deviation from the gene's mean expression across all experiments. In this data, we treat each gene as an instance that is described by its behavior in the different experiments. We randomly partitioned the data into 4922 training instances (genes) and 1230 test instances.

Table 1: Comparison of the IB-EM algorithm, 50 runs of EM with random starting points and 50 runs of mean field EM from the same random starting points. Shown are train and test log-likelihood per instance for the best and 80th percentile of the random runs. Also shown is the percentile of the runs that are worse than the IB-EM results. Datasets shown include a Naïve Bayes model for the Stock dataset, and the Digit dataset; a 3 and 4 level hierarchical model for the Digit dataset (DigH3 and DigH4); and an hierarchical model for the Yeast dataset. For each model we show several cardinalities for the hidden variables, shown in the first column.

| Model | Train Log-Likelihood | | | | | | | Test Log-Loss | | | | | | |
| | IB-EM | Restarts EM | | | Mean Field EM | | | IB-EM | Restarts EM | | | Mean Field EM | | |
| | | %< | 100% | 80% | %< | 100% | 80% | | %< | 100% | 80% | %< | 100% | 80% |
| **Stock** | | | | | | | | | | | | | | |
| C=3 | -19.91 | 62% | -19.90 | -19.90 | | | | -19.90 | 76% | -19.88 | -19.89 | | | |
| C=4 | -19.47 | 98% | -19.46 | -19.52 | | | | -19.52 | 96% | -19.52 | -19.62 | | | |
| C=5 | -19.16 | 94% | -19.15 | -19.24 | | | | -19.31 | 98% | -19.30 | -19.39 | | | |
| **Digit** | | | | | | | | | | | | | | |
| C=5 | -297.42 | 100% | -299.16 | -306.47 | | | | -319.48 | 100% | -320.74 | -327.42 | | | |
| C=10 | -269.54 | 100% | -287.50 | -295.45 | | | | -333.29 | 76% | -326.10 | -332.52 | | | |
| **DigH3** | | | | | | | | | | | | | | |
| C=2 | -333.05 | 94% | -332.69 | -333.59 | 100% | -333.41 | -334.12 | -325.176 | 94% | -331.32 | -333.08 | 100% | -332.68 | -333.72 |
| C=3 | -313.66 | 98% | -313.17 | -315.29 | 100% | -315.32 | -316.53 | -321.617 | 98% | -320.80 | -323.33 | 100% | -322.95 | -324.47 |
| C=4 | -299.21 | 100% | -301.18 | -304.20 | 100% | -302.84 | -306.40 | -312.979 | 100% | -317.79 | -320.29 | 100% | -318.97 | -322.76 |
| **DigH4** | | | | | | | | | | | | | | |
| C=2 | -322.77 | 20% | -320.45 | -321.07 | 100% | -323.11 | -323.98 | -325.276 | 30% | -322.67 | -323.41 | 100% | -326.11 | -326.82 |
| C=3 | -297.87 | 14% | -294.22 | -295.56 | 100% | -299.59 | -301.02 | -309.991 | 12% | -304.93 | -306.83 | 100% | -311.74 | -312.86 |
| C=4 | -284.42 | 8% | -278.80 | -281.03 | 100% | -287.80 | -288.41 | -303.602 | 8% | -297.31 | -299.65 | 100% | -305.45 | -308.07 |
| **Yeast** | | | | | | | | | | | | | | |
| C=2 | -149.80 | 22% | -148.33 | -148.66 | 100% | -150.01 | -150.35 | -148.89 | 28% | -147.48 | -147.78 | 100% | -149.12 | -149.52 |
| C=3 | -141.72 | 0% | -139.58 | -139.77 | 100% | -141.84 | -142.07 | -140.78 | 0% | -138.56 | -138.76 | 100% | -140.91 | -141.13 |
| C=4 | -139.60 | 0% | -136.48 | -136.66 | 100% | -139.65 | -139.80 | -138.93 | 0% | -135.68 | -135.87 | 92% | -138.82 | -139.00 |

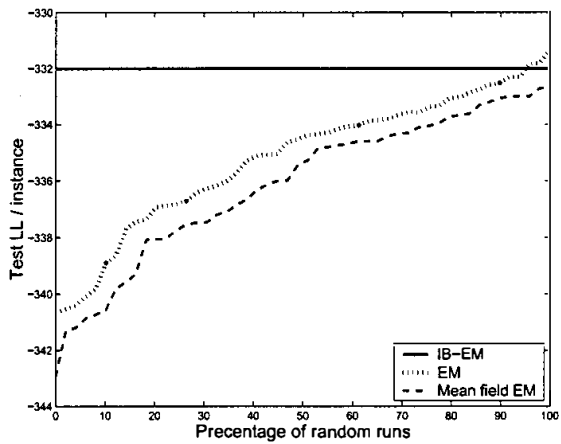

Figure 4: Comparison of test performance of the **IB-EM** algorithm to the cumulative performance of 50 random EM and mean field EM runs on the 3-level hierarchy model with binary variables for the Digit domain.

The model we learned for this data has an hierarchical structure with 19 hidden variables in a 3-level hierarchy that was determined by the nature of the different experiments. The middle level has a hidden parent to all similar conditions (e.g., different types of heat shock), and the top level contains a single variable that is parent of all variables in the middle level.

As a first sanity check, for each model (and each cardinality of hidden variables) we performed 50 runs of EM with random starting points. These resulted in parameters that have a wide range of likelihoods both on the training set and the test set. These results (which we elaborate on below), indicate that these learning problems are challenging in the sense that EM runs are trapped in different local maxima.

Next, we considered the application of IB-EM on these problems. We performed a single IB-EM run on each problem and compared it to the 50 random EM runs, and also to 50 random mean field EM runs. For example, Figure 4 compares the test set performance (log-likelihood per instance) of these runs on the Digit dataset with a 3-level hierarchy of binary hidden variables. The solid lines shows the performance of the IB-EM solution at $\gamma = 1$. The two dotted lines show the accumulative performance of the random runs. As we can see, the IB-EM model is superior to all the mean field EM runs, and to 94% of the EM runs.

It is important to note the time required by these runs, all on a Pentium IV 2.4 Ghz machine. A single mean field EM run requires approximately 1 minute, an exact EM random run requires roughly 25 minutes, and the single IB-EM run took 27 minutes. Thus, IB-EM takes about the same time as 25-30 mean field EM runs but reaches solutions that none of these runs achieve. The IB-EM run time is roughly similar to the exact EM in terms of time, but we would need about 20 such runs to get to a comparable solution. These proportions hold for other datasets, although for harder problems the standard EM runs are somewhat

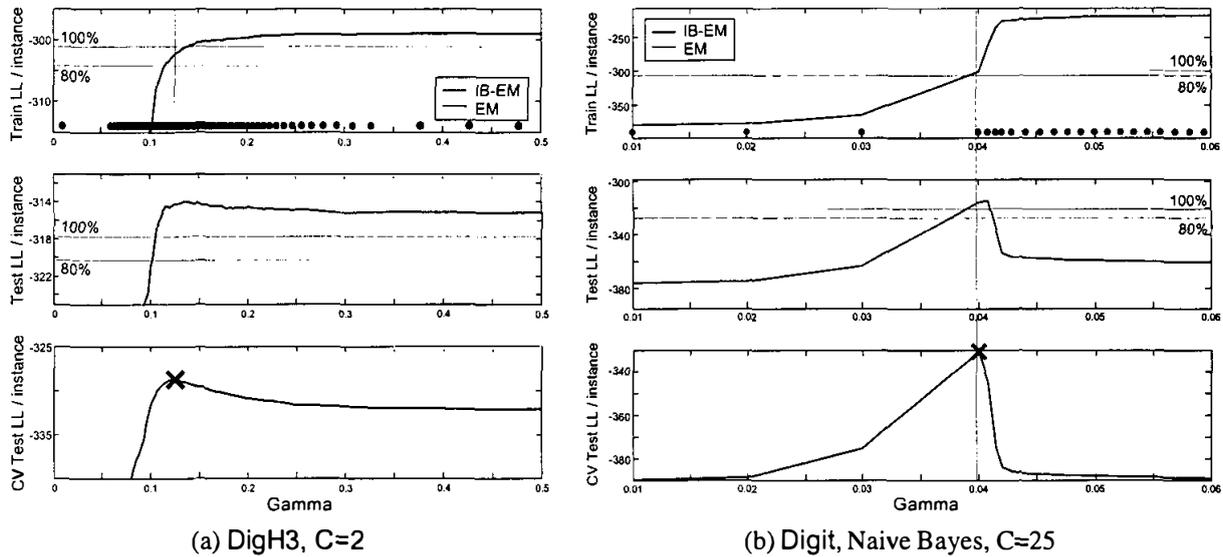

(a) DigH3, C=2                                              (b) Digit, Naive Bayes, C=25

Figure 3: Continuation performance for two runs. (top panel) Log-likelihood per instance vs. $\gamma$. The dotted lines show the best and 80th percentile of 50 random EM runs. (middle panel) Log-likelihood of unseen test data. (bottom panel) Predicted test performance using cross validation on training data. Vertical line, denotes CV estimate of $\gamma^*$. Circles mark values of $\gamma$ for which the Lagrangian was evaluated during the continuation process.

more expensive (4-5 times) than the IB-EM runs.

Table 1 summarizes the results of these comparisons on the different learning problems. It shows both the training set and test performance of the different methods. We show the best and the 80% run for each group of 50 random starting points, as well as the relative percentile of the IB-EM solution among these runs. As we can see, IB-EM runs outperforms random restarts for the Naive Bayes models in most cases. For the harder problems that involve a hierarchy of hidden variables, the situation is more complex. When we compare IB-EM to the mean field EM, we see that it is consistently better, often by a non-trivial margin.

When we compare IB-EM to exact EM we see that in the simpler 3-level Digit hierarchal models IB-EM is better than most of the EM runs. However, in the more complex hierarchal models the performance of IB-EM is worse than most EM runs. Our hypothesis is that this drop in performance is due to the inherent limitations of the mean field approximation in these models. This approximation loses much of the information about interactions between the hidden variables. We stress, however, that we selected models where we can perform exact inference over the hidden variables so that we can compare to exact EM. In many applications exact inference is infeasible, and approximations are needed. Clearly the mean field approximation is quite crude. Yet, as we discussed above, our framework allows to use more refined variational approximations, and we expect that these will improve the performance for both variational EM and IB-EM.

We also compared the IB-EM method to the perturbation method of Elidan et al [4]. Briefly, their method alters the landscape of the likelihood by perturbing the relative weight of the samples and progressively diminishing this perturbation as a factor of the temperature parameter. In the **Stock** dataset, the perturbation method initialized with a starting temperature of 4 and cooling factor of 0.95, had performance similar to that of IB-EM. However, running time of the perturbation method was an order of magnitude larger. For other examples we considered above, running the perturbation method with the same parameters proved to be prohibitively expensive. When run with more efficient parameter settings, the perturbation method's performance was inferior to that of IB-EM. These results are consistent with those of Elidan et al [4] who showed some improvement for the case of parameter learning but mainly focused on structure learning, with and without hidden variables.

We now turn to examine the continuation process more closely. Figure 3(a) illustrates the progression of **IB-EM** (on DigH3, C=2). The top panel shows training log-likelihood per instance of parameters in intermediate points in the process. This panel also shows the values of $\gamma$ evaluated during the continuation process (circles). As we can see, the continuation procedure focuses on the region where there are significant changes in the likelihood. The middle panel shows the likelihood on test data. As we can see, although the training set likelihood increases as we increase $\gamma$, the test set performance deteriorates at some stage, due to overfitting the data.

This example suggests that we can improve performance by using the model at $\gamma^*$ instead of at $\gamma = 1$. In section 4.4 we suggested using a CV test that is based *only* on the training data to estimate the critical value $\gamma^*$. The lower panel shows the CV estimate of the test set likelihood using only the training data. Although these estimates are biased, we see that they allow the learning algorithm to pinpoint the value of $\gamma^*$. To demonstrate the effectiveness of this approach more clearly we examine a situation where we have an excessive number of parameters. Namely, the Naive Bayes model on the Digit dataset with $C = 25$. In this scenario, we expect the learned model to overfit the training data. And thus, we expect that $\gamma^*$ is lower then 1. Indeed, as we can see in Figure 3(b) at early stages in the learning the model clearly overfits the data. However, using CV estimate of of $\gamma^*$, the procedure learns a model with test set performance of $-315.78$ which is much better than all other Naive Bayes models learned on this dataset.

## 7 Discussion and Future Work

In this work we set out to learn models with hidden variables. We described a method for reaching a high-scoring solution by starting with a simple solution and following a trajectory to a high-scoring one. The contribution of this work is threefold.

First, we made a formal connection between the Information Bottleneck principle [17, 6] and maximum likelihood learning for graphical models. The Information Bottleneck and its extensions are originally viewed as methods to understand the structure of a distribution. We showed that in some sense the Information Bottleneck and maximum likelihood estimation are two sides of the same coin. The Information Bottleneck focuses on the distribution of variables in each instance, while maximum likelihood focuses on the projection of this distribution on the estimated model. This understanding extends to general Bayesian networks the recent results of Slonim and Weiss [16] that relate the original Information Bottleneck and maximum likelihood estimation in univariate mixture distributions.

Second, the introduction of the IB-EM principle, allowed us to use an approach that starts with a solution at $\gamma = 0$ and progresses toward a solution in the more complex landscape of $\gamma = 1$. This general scheme is common in *deterministic annealing* approaches [15, 18]. These approaches "flatten" the landscape by raising the likelihood to the power of $\gamma$. The main technical difference of our approach is the introduction of a regularization term that is derived from the structure of the approximation of the probability of the latent variables in each instance.

Third, we applied continuation methods for traversing the path from the trivial solution at $\gamma = 0$ to a solution at $\gamma = 1$. Unlike standard approaches in deterministic annealing and Information Bottleneck, our procedure can automatically detect important regions where the solution changes drastically and ensure that they are tracked closely. In preliminary experiment results (not shown) the continuation method was clearly superior to standard annealing strategies. As we show in our experimental results, applying IB-EM to hard learning problems leads to solutions that are often superior to the equivalent version of EM. More-over, by using an early stopping rule, we can find solutions at intermediate values of $\gamma$ that are better at generalization.

The methods presented here can be extended in several directions. First, by relaxing the mean field variational approximation, we can explore a better tradeoff between tractability and quality of learned model. Second, we have used an approximate solution for performing continuation. Better approximations might lead to more accurate results (with fewer steps). Third, we showed that by using cross validation we can detect $\gamma^*$ values where generalization is better. Deriving a principled method for understanding this generalization has both theoretical implications and can lead to faster and more accurate learning. Finally, the same principle can be generalized to problems of *structure* learning, by replacing the parameter optimization step with a structure learning procedure. This results in a natural extension of the *structural EM* [5] framework for learning the structure as well as parameters of the Bayesian network of interest.

**Acknowledgements**

We thank R. Bachrach, Y. Barash, G. Chechik, D. Koller, M. Ninio, T. Tishby, and Y. Weiss for discussions and comments on earlier drafts of this paper. This work was supported, in part, by a grant from the Israeli Ministry of Science. G. Elidan was also supported by the Horowitz fellowship. N. Friedman was also supported by an Alon Fellowship and by the Harry & Abe Sherman Senior Lectureship in Computer Science..